*Article*

# Zero-shot sketch-based remote sensing image retrieval based on multi-level and attention-guided tokenization

Bo Yang [1,2], Chen Wang [1,2]*, Xiaoshuang Ma [1,2]*, Beiping Song [1,2], Zhuang Liu [3] and Fangde Sun [4]

[1] Anhui Province Key Laboratory of Wetland Ecosystem Protection and Restoration, Anhui University, Hefei 230601, China;
[2] School of Resources and Environmental Engineering, Anhui University, Hefei 230601, China;
[3] Shanghai Ubiquitous Navigation Technology Co. Ltd, Shanghai 201702, China;
[4] The 54th Research Institute of China Electronics Technology Group Corporation, Shijiazhuang 050081, China;
* Correspondence: chen.wang@ahu.edu.cn; mxs.88@whu.edu.cn

**Abstract:** Effectively and efficiently retrieving images from remote sensing databases is a critical challenge in the realm of remote sensing big data. Utilizing hand-drawn sketches as retrieval inputs offers intuitive and user-friendly advantages, yet the potential of multi-level feature integration from sketches remains underexplored, leading to suboptimal retrieval performance. To address this gap, our study introduces a novel zero-shot, sketch-based retrieval method for remote sensing images, leveraging multi-level feature extraction, self-attention-guided tokenization and filtering, and cross-modality attention update. This approach employs only vision information and does not require semantic knowledge concerning the sketch and image. It starts by employing multi-level self-attention guided feature extraction to tokenize the query sketches, as well as self-attention feature extraction to tokenize the candidate images. It then employs cross-attention mechanisms to establish token correspondence between these two modalities, facilitating the computation of sketch-to-image similarity. Our method significantly outperforms existing sketch-based remote sensing image retrieval techniques, as evidenced by tests on multiple datasets. Notably, it also exhibits robust zero-shot learning capabilities in handling unseen categories and strong domain adaptation capabilities in handling unseen novel remote sensing data. The method's scalability can be further enhanced by the pre-calculation of retrieval tokens for all candidate images in a database. This research underscores the significant potential of multi-level, attention-guided tokenization in cross-modal remote sensing image retrieval. For broader accessibility and research facilitation, we have made the code and dataset used in this study publicly available online. Code and dataset are available at https://github.com/Snowstormfly/Cross-modal-retrieval-MLAGT.

**Keywords:** remote sensing image retrieval; zero-shot learning; attention mechanism; deep learning; transformers

## 1. Introduction

The proliferation of remote sensing sensors deployed on various carrier platforms has led to a continuous and rapid increase in the volume of observation data pertaining to the Earth's surface. The data generated by these sensors possesses the characteristics commonly associated with big data: it is voluminous, exhibits a wide variety, is generated at high velocity, and requires rigorous verification. While this abundance of data offers users unprecedented opportunities to discover and quantify underlying phenomena, it also presents many challenges [1-3]. The challenges in remote sensing big data include data management, analysis, retrieval, and interpretation complexities. Among these challenges, data retrieval—accurately and efficiently finding the desired category images from a massive amount of remote sensing images, is crucial for subsequent data mining processes[4]. Traditional query inputs typically involve image metadata values, such as bounding box coordinates, sensor names, and timestamps. Another commonly employed approach, which is content-based remote sensing image retrieval (CBRSIR) involves querying remote sensing warehouses using example images, often yielding promising results when coupled with state-of-the-art deep learning algorithms [5,6]. However, in numerous application scenarios, users encounter difficulty in providing desirable remote sensing examples. Consequently, some cross-model retrieval method, such as text-image retrieval [7] and sketch-based remote sensing image retrieval (SBRSIR) [8] has captured the attention of some researchers. As demonstrated in Figure 1. SBRSIR offers users the ability to express the structure of a desired remote sensing image from their mind through freehand sketches, which can then be employed as queries for retrieving images. This method is believed to be intuitive for users, easy to execute on touch-enabled devices, and capable of achieving a high level of expressiveness and flexibility [9-12]. From a pragmatic perspective, our research team has been working on a project known as "Habitat Yangtze", which is one part of the broader Space Climate Observatory initiative (visit www.spaceclimateobservatory.org/habitat-yangtze for more information). The objective of this project is to offer sophisticated remote sensing and mapping services to varied users, including wetland managers, bird watchers, and climate change researchers, many of whom possess limited expertise in remote sensing technologies. These stakeholders have demonstrated a significant interest in a sketch-based remote sensing image retrieval system, highlighting the limitations of traditional query inputs in representing their visions and imaginations. This expressed need has significantly inspired and directed the focus of our research.

Sketch-based image retrieval (SBIR) for common images have undergone extensive investigation in recent years and yielded promising results [13-17]. To bridge the two modalities, a common approach contains three stages: feature extraction from both modalities, feature enhancement, and image retrieval. The latest solutions, like ACNet [18], DAL [19], and ZSE-SBIR [13], often employ deep structures like Vit (Vision transformer), Resnet, and use homogeneous [20], Siamese branch [9], or heterogeneous structures [21] for different modalities. Despite the inspiring and promising development in SBIR, only a small number of recent research publications have dedicated to Remote sensing SBIR [8,22-26]. Current SBRSIR models underscores the imperative for novel strategies that can surpass the existing limitations, especially in terms of retrieval accuracy, zero-shot learning capability, and domain adaptation capability. Compared to ordinary images, remote sensing images are often obtained from aerial viewpoints and covers a much larger geographical area. This leads to noticeable semantic and structural disparities with ordinary images. This constitutes the principal difference between SBIR and SBRSIR research and makes the direct application of SBIR questionable. Also, the benchmark training and testing datasets for common pictures are no longer pertinent for remote sensing. Consequently, researchers in the field of remote sensing applications often find themselves compelled to design and train ad-hoc models. There are two primary challenges in the domain of SBRSIR. First, there is an extensive array of categories concerning objects and scenes within remote sensing images. The sketch samples are largely scarce, compared with the ordinary photo-related sketches [8]. As a result, constructing a comprehensive training dataset for remote sensing SBIR proves to be a formidable task. In contrast to the general image processing domain, well-annotated datasets for remote sensing images and sketches remain insufficient for supervised learning [23], thus, potentially affecting the performance of trained networks, particularly when confronted with unseen categories and images. Given the challenges associated with significantly expanding training datasets to encompass all potential categories and data sources in remote sensing, existing SBRSIR research underscores the importance of zero-shot learning performance [23,24]. Nevertheless, the existing models in this domain still exhibit very limited zero-shot capabilities. The second challenge is that remote sensing images exhibit distinct characteristics depending on the sensor and their carrier platform. The SBRSIR model should have a good domain adaptation capability to support different sources of remote sensing images. However, current work investigated very little in the domain adaptation capability of the SBRSIR model.

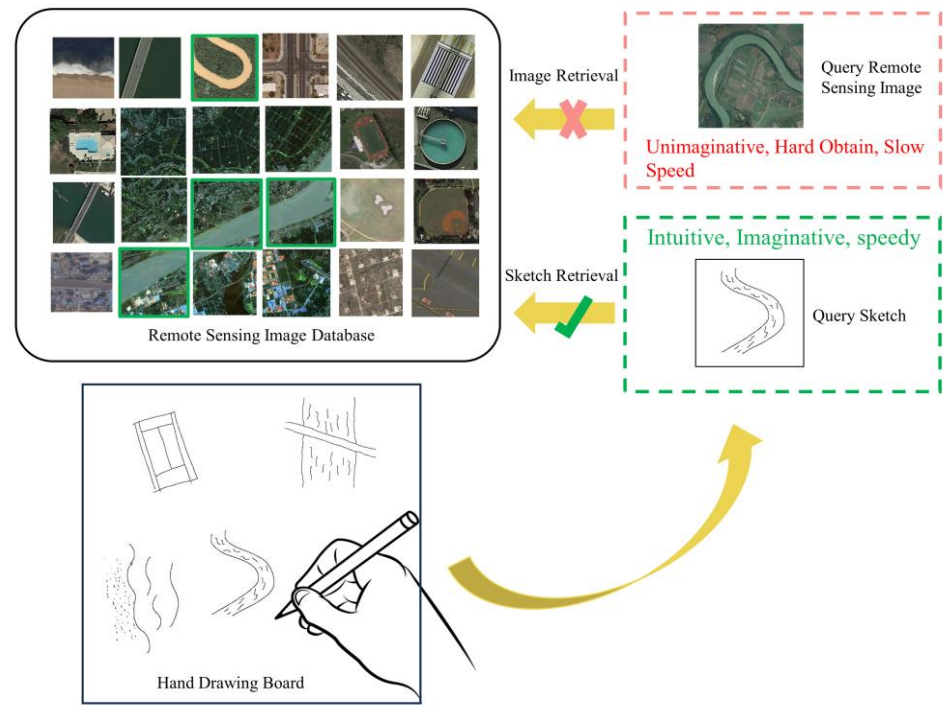

**Figure 1.** Demo of sketch-based remote sensing image retrieval process and its advantages compared to content-based retrieval.

In response to the growing demand and challenges of SBRSIR, we propose a novel zero-shot cross-model deep learning network that leverages a new multi-level feature extraction and attention-guided tokenization mechanism. Additionally, we have substantially expanded the RSketch SBRSIR benchmark dataset to serve as a comprehensive testbed, allowing for a comprehensive evaluation of the performance of our new method in various settings. Our test results reveal that our new method significantly outperforms existing SBRSIR algorithms. Our proposed algorithm exhibits excellent zero-shot capabilities, enabling precise retrieval of images from unseen categories, and demonstrates strong domain adaptation capabilities, as it can be trained on mixed-source remote sensing image samples and retrieve images from previously unseen data sources, thereby offering extensive application flexibility.

The main contributions and innovations of our proposed method can be summarized in two key aspects:

(1) We introduced a novel deep learning categorical SBRSIR network equipped with a multi-level feature extraction, self-attention-guided tokenization and filtering, and cross-modality attention update across sketch and remote sensing modalities. This approach simplifies the comparison process to focus on a select number of significant patches from remote sensing images relative to the query sketch. Moreover, our model eliminates the requirement for semantic knowledge input, significantly lowering the costs associated with constructing training datasets. This new approach has demonstrated substantial performance improvements compared to existing methods, particularly in terms of its good zero-shot capability. Furthermore, our proposed model showcases impressive domain adaptation capabilities as it can be trained using mixed-source remote sensing datasets and can retrieve remote sensing image from unseen sources.

(2) We have substantially expanded SBRSIR benchmark RSketch dataset to Rsketch_Ext dataset as testbed. This new testbed encompasses 20 categories, each containing 90 sketches and over 400 remote sensing images from various datasets. We have made this comprehensive dataset as well as the code of our method available online to facilitate the work of researchers in this field.

The rest of this article is organized as follows. Section 2 describes the related work of SBRSIR. Section 3 gives the details of our proposed method. Some experimental results are reported in Section 4. The discussion is given in Section 5 and conclusion is drawn in Section 6.

## 2. Related works

This section explores prior research in the field of SBIR, SBRSIR, and other application domains within image retrieval. The utilization of freehand-drawn sketches as input queries has attracted researchers in computer vision [27,28], human-computer interaction [29], and geographical information science [30] since its early stages. The inherent challenge in sketch-based retrieval lies in the multi-modality nature of such a task. Sketches significantly differ from photographic or remote sensing images due to their abstract, symbolic, sparse, and stylistic nature, often containing rich topological and semantic information.

SBIR can be categorized into two levels in terms of granularity: categorical [31] and fine-grained [32]. Categorical SBIR focuses on recognizing the categorical information in sketches, retrieving images from the same category, often described by semantic notions like "plan". In remote sensing domain, the development is currently at a categorical level [8,23]. On the other hand, fine-grained or instance-level SBIR delves into scene and stroke details, such as relative location and topology, for more precise image retrieval. Both categorical and fine-grained SBIR may frequently encounter images from unseen categories during testing. To address this, Zero-Shot Learning (ZSL) algorithms are introduced, often employing assisting information such as word embeddings for semantic similarity measurement [23,33,34]. However, incorporating semantic information directly can present significant challenges in constructing training datasets and may also restrict the model's generalizability. Recent works aim to enhance algorithm's generalizability, adopting features and local correspondence information for fine-grained yet generalizable algorithms [13,35]. The improvement of algorithms' generalizability may also be helpful in cross-dataset retrieval [13]. Despite advancements in computer vision domain, such algorithms have not been explored in the context of sketch-based retrieval in the remote sensing domain.

### 2.1. Cross-Model Feature Extraction

Most contemporary SBIR and SBRSIR algorithms generally follow a three-part pipeline: cross-model feature extraction, feature enhancement, and image retrieval. The feature extracts part extract informative and representative features from both sketches and images of other modalities. Initially, hand-crafted features like gradient field HOG [27], edge maps [36], and histogram of edge local orientations [37] were applied. However, recent years have seen the widespread adoption of deep learning [9,38] in SBIR and SBRSIR [25,26], outperforming handcrafted methods [39]. Various deep network structures, including FCN [40], CNN [41], RNN [28], VAE [42], GNN [43], transformers [44], and ViT [13], have been explored. The latest solutions often combine multiple deep structures [13,41] and employ homogeneous [20], Siamese branch [9], or heterogeneous structures [21] for different modalities.

### 2.2. Feature Enhancement

Feature enhancement part involves feature selection, aggregation, and embedding [16]. This step prepares the extracted feature for comparison and retrieval. Recent content-based image retrieval research [45-47] utilizes attention maps to weigh feature importance, with [48] further proposing feature elimination or merging for efficiency without significant accuracy loss—a novel approach in SBIR. For feature embedding, common methods include Bag of features [49], Bag of visual words [27], VLAD [50], and FV [51]. To enhance scalability, deep features could be pre-calculated and embedded in binary hash codes [52,53].

*2.3. Image Retrieval*

After encoding features in the embedding space, image retrieval typically involves a Nearest Neighbor search based on global similarity calculated through Euclidean distance [25] or Hamming distance for binary hash codes [54]. Deep metric learning is also prevalent in similarity calculation in SBIR [55,56]. Instead of global similarity calculation, some works focus on local matching pairs between sketches and images for similarity scoring, such as [57] using the quantity of matching pairs and [13] employing a cosine similarity matrix. Approaches like Deep Hashing [52], Approximate Nearest Neighbor (ANN) using k-means tree [58], and ANN using product-quantization [52] demonstrate scalability for efficient retrieval from large image databases.

*2.4. Training and Data*

Supervised and unsupervised deep learning approaches are both employed in SBIR [52,59-61], with most SBRSIR research using supervised training. In supervised training process, adequately annotated sketch-remote sensing datasets are crucial. However, there are only two dedicated SBRSIR benchmark datasets identified—RSketch [8] and Earth on Canvas [23]. They are relatively small compared to standard SBIR datasets like Quickdraw extend [28] and sketchy [54]. Also, these two datasets only cover a small fraction of all potential sensors. The scarcity of large training data emphasizes the significance of zero-shot, semantic knowledge independent, and domain adaptation capabilities in SBRSIR—a key improvement in our proposed method.

In sum, compared to SBIR, research on SBRSIR remains limited and is mostly focused on category-based retrieval. The lack of comprehensive training and testing benchmark datasets means that zero-shot learning and domain adaptation capabilities are critical, yet the capabilities of existing models are not ideal. Research in SBRSIR urgently requires the development of new algorithms with better zero-shot learning and domain adaptation capabilities.

**3. Methodology**

Table 1 displays the symbols used in this chapter along with their corresponding.

| Symbol | Interpretations |
|---|---|
| S | Query Sketch |
| R | Remote sensing image |
| $D^R$ | Remote sensing image database |
| E | Visual token embedding |
| [RT] | Retrieval token |
| Q | Queries in transformers |
| K | Keys in transformers |
| V | Values in transformers |

Informed by the fundamental principles delineated in prior studies [13,62,63], the proposed model in this research consists of two stages: a self-attention for feature extraction and a cross-attention for similarity calculation. The structural design of this model is visually represented in Figure 2, which describes the model's capacity for facilitating the retrieval process across two distinct modalities: remote sensing imagery and sketches. Detailed explications of each stage's specific functions and their role within the model are systematically presented in the following sections.

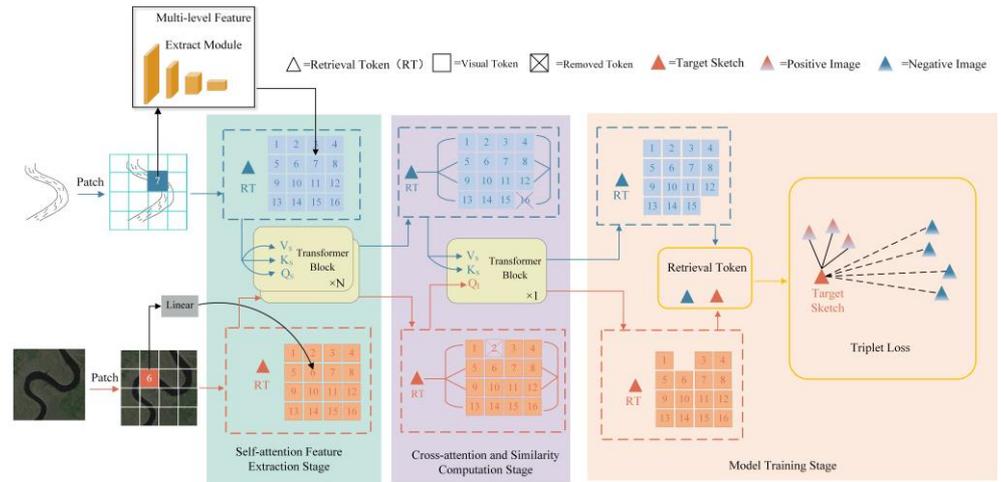

**Figure 2.** Model overview: (1) Self-attention feature extraction stage, multi-level extraction and filtering of feature information for different modes. (2) Cross-attention and similarity computation stage, establishing the correspondence of two modes and calculating the similarity (3) Model training stage, train the network using triplet loss.

### 3.1. Self-attention Feature Extraction

In the context of sketch-based remote sensing image retrieval, the process typically begins with a query sketch *S*. The objective is to identify the corresponding remote sensing image *R* from the remote sensing image database $D^R$. Traditional deep network methodologies transform hand-drawn sketches and remote sensing images into a sequence of visual tokens. These methods predominantly capture local features of sketches, which are often characterized by sparse strokes, as noted in [13]. To address this limitation and effectively expand the receptive field of visual tokens to better accommodate sparse strokes, this study implements a multi-level feature extraction module applicable to sketches. This enhancement aims to preserve a more comprehensive set of feature information from sketches. The module is constructed by layering multiple convolutional layers with distinct kernel sizes and includes learnable parameters. Following each convolutional layer is a non-linear activation function. This configuration results in an $n \times d$ dimensional visual token embedding $E = [E^1, E^2, \ldots, E^n]$, where each *E* represents a *d*-dimensional vector.

This module incorporates a transformer-based self-attention mechanism, an approach inspired by the human visual and cognitive systems. This mechanism enables neural networks to dynamically concentrate on the most informative segments within the input data. By integrating the self-attention mechanism, the neural networks in our model are designed to autonomously identify and focus on salient features in the input, whether it be a sketch or a remote sensing image. This strategy can significantly enhance the network's performance and its ability to generalize.

The initial step in our methodology involves processing the output from the convolutional layers. This output is first concatenated with a retrieval token *[RT]*, a trainable *d*-dimensional embedding vector that embodies the global feature. This combination results in an augmented *n*+1 dimensional visual token embedding $E' = [RT, E^1, E^2, \ldots, E^n]$. Then, the self-attention is realized by passing $E'$ through first the Multi-Head Self Attention (MSA) module and second the Multi-Layer Perceptron (MLP) module. The forward propagation formula of the model is as follows:

$$E_0 = E' \tag{1}$$

$$E_l = MSA\big(LN(E_{l-1})\big) + E_{l-1}, \quad l = 1 \ldots L \tag{2}$$

$$E_{l'} = MLP(LN(E_l)) + E_l, \ l = 1 \dots L \tag{3}$$

Formulas (2) and (3) both incorporate residual connections, where $L$ represents the number of hidden attention layers and $LN$ stands for layer normalization.

The MSA module is a core component of our deep network, designed to discern interrelations among various token vectors within a remote sensing image or a sketch. The essential element of this module is the scaled dot-product attention mechanism. Initially, the layer-normalized visual token embedding is multiplied by three learnable matrices $W_q$, $W_k$ and $W_v$, resulting in $Q$ (Queries), $K$ (Keys), and $V$ (Values) per the following equations:

$$Q = E \cdot W_q \quad K = E \cdot W_k \quad V = E \cdot W_v \tag{4}$$

Then the scaled dot-product attention is calculated by the following equation:

$$Attention_{self}(Q, K, V) = softmax(\frac{QK^T}{\sqrt{d}})V \tag{5}$$

In this equation, the product of $Q$ and $K$ assesses the similarity between the Query and Key. The result is then scaled by the square root of the dimension $d$ of $E^i$ to mitigate vanishing gradient problem. A softmax function normalizes the similarities across multiple Keys relative to a Query, ensuring their cumulative sum equals 1. The resulting similarity is used as weights to compute the weighted average of the corresponding $V$, ultimately obtaining an attention head of $(n + 1) \times d$-dimension. This procedure is replicated $h$ times to create $h$ attention heads. These heads are subsequently integrated into $(n + 1) \times d$-dimensional MSA output by a dense network. The output of MSA is then followed by a MLP module, as described in [62].

The proposed method integrates feature filtering at specific layers of the self-attention stage. Given the varying degrees of information richness among local visual tokens generated through self-attention, selectively filtering out tokens with lesser feature information can reduce the number of tokens and improve efficiency for both training and referencing. The filtering is achieved by leveraging attention scores between *[RT]* and all other visual token embedding vectors. Specifically, the typical Query of visual token embedding is replaced with the Query of *[RT]*. The formula is as follows:

$$Attention_{filtering}(Q, K, V) = sofamax(\frac{Q_{[RT]}K^T}{\sqrt{d}}) \tag{6}$$

Utilizing this equation enables the computation of attention scores between *[RT]* and all visual tokens. Based on the attention scores, only $k$ visual token vectors are retained for further processing. Consequently, this leads to a more refined set of visual token embeddings for both the sketch image and remote sensing images, denoted as $E'' = [RT_{final}, E^1_{final}, \dots, E^k_{final}]$ (where $k < n$).

*3.2. Cross-attention and Similarity Calculation*

Our method employs cross-attention to establish cross-modal token embedding correspondences between sketches and remote sensing images. This involves an interchange of the sketch query $Q_S$ and the candidate remote sensing image query $Q_R$. After swap, the Query, Key, and Value for the sketch and remote sensing image become $(Q_R, K_S, V_S)$ and $(Q_S, K_R, V_R)$ respectively. This interchange facilitates a direct connection between the visual token embedding sets of the sketches and the remote sensing images. Taking $Q_S$ as an example, the cross-modal attention is obtained using the following formula:

$$Attention_{cross}(Q_S, K_R, V_R) = softmax(\frac{Q_S K_R^T}{\sqrt{d}})V_R \tag{7}$$

Through this attention mechanism, the visual token embeddings of both the sketch and the remote sensing image, including the retrieval token *[RT]*, are updated based on the pair-wise token information from each modality.

The final step in our methodology involves the use of Euclidean distance between $[RT]_S$ and $[RT]_R$ as a metric for measuring the similarity between the sketch input and a candidate remote sensing image.

*3.3. Model Training and Image Retrieval*

In this study, the deep network is trained using the triplet loss function with *[RT]* derived from both sketches and remote sensing images. Specifically, we consider a triplet $(S_i, R_i^+, R_i^-)$ in the training set, where $S_i$ denotes a query sketch, $R_i^+$ denotes a remote sensing image with the same label as $S_i$, and $R_i^-$ denotes a remote sensing image with a different label. The primary aim of this loss function is to minimize the distance between correctly matched sketch-remote sensing image pairs (positive examples) while ensuring that the distance between each sketch and incorrectly matched remote sensing images (negative examples) exceeds a predefined margin. Here, *[RT]* is used as the global descriptor for sketches and remote sensing images. The triplet loss is defined as the following equation:

$$L_{tri} = \frac{1}{T}\sum_{i=1}^{T} max\left\{\left\|[RT]_{S_i}, [RT]_{R_{i+}}\right\|_2 - \left\|[RT]_{S_i}, [RT]_{R_{i-}}\right\|_2 + m,\ 0\right\} \quad (8)$$

In this equation, *T* represents the total number of triplets, and *m* denotes the margin that discriminates whether a sketch and a remote sensing image are from the same class. If a sketch and a candidate remote sensing image's difference exceeds the margin value, they are not in the same class.

In the retrieval phase, the query sketch is processed through the network to obtain its retrieval token $[RT]_S$. Similarly, each candidate remote sensing images in the database are transformed into $[RT]_R$. The similarity between *[RT]* of the sketch input and a candidate remote sensing image is measured by calculating their Euclidean distance. A smaller Euclidean distance indicates a closer similarity. Subsequently, the remote sensing images that exhibit the smallest distances from the query sketch—essentially, its k-nearest neighbors—are selected as the final output of the model's retrieval process.

To further improve the response speed during actual retrieval operations, $[RT]_R$ is pre-computed and stored in the database by invoking the model in advance. During retrieval, the model is only required to compute the retrieval token for the input sketch. Following this, the pre-stored $[RT]_R$ from the remote sensing image database are rapidly accessed, and the Euclidean distance between $[RT]_S$ and all $[RT]_R$ in the database can be directly calculated. This approach significantly accelerates the retrieval process, especially when dealing with big remote sensing image collections. Moreover, the efficiency of the retrieval process can be further augmented by employing techniques such as approximate nearest neighbor algorithms or utilizing a vector database. This improvement is particularly useful as the volume of images in the remote sensing database expands, underscoring the practical scalability and efficiency of the model in real-world retrieval scenarios. These mechanisms are crucial for optimizing the model's performance, particularly in retrieval applications where speed and accuracy are paramount.

**4. Experiments**

*4.1. Dataset*

The development of effective cross-modal retrieval models, particularly for SBRSIR, is hindered by the scarcity of specialized datasets. While there are numerous cross-modal retrieval datasets for natural images, such as Sketchy, TUBerkin, and QuickDraw, the availability of similar datasets for remote sensing images is very limited. This gap presents a significant challenge for training SBRSIR models.

This study used RSketch dataset [8], RSketch_Ext dataset (expanded based on RSketch dataset), Earth on canvas dataset [23], UCMerge Landuse dataset [64], and GF-1 image tiles of the Anhui Province section in the middle and lower basin of the Yangtze River. The RSketch dataset, a publicly available sketch-remote sensing image dataset, comprises 20 categories including airplanes, baseball fields, and bridges, each with 45 sketches and 200 remote sensing images. In this study, we expanded the RSketch dataset to RSketch_Ext by augmenting each category with additional remote sensing images sourced from various public datasets like AID [65], NWPU-RESISC45 [66], WHU_RS19 [67] and others. Furthermore, as part of our HABITAT YANGTZE project, we enlisted the help of 10 non-professional (in remote sensing) volunteers, none of whom possess a background in drawing, to engage in a sketching task. Each volunteer was instructed to create sketches in five categories based on their personal interpretation of the shapes associated with those categories. We then collected the sketches produced by the volunteers and obtained a specific number of sketches from each completed category to enrich the diversity of our augmented dataset's sketch categories. Concurrently, we enhanced our sketch dataset further by producing simulate sketch from OpenStreetMap (OSM) data. We transformed OSM data into sketch-like images, and integrated these into appropriate sketch categories of our dataset. Post-expansion, each category includes 90 sketches and at least 400 remote sensing images, significantly increasing the dataset's size. The sketches are in both TIFF and JPEG formats. Figure 3 presents sample data from each category in the RSketch_Ext dataset. The Earth on Canvas dataset contains 14 categories, five of which are unique compared to those in the RSketch and RSketch_Ext dataset, with each category comprising 100 sketches and 100 remote sensing images. The UCMerge Landuse dataset is a publicly available dataset for remote sensing image classification. It features a rich set of image categories in comparison to other remote sensing image datasets, comprising a total of 21 categories. Among these, 10 categories align with those in the RSketch dataset and RSketch_Ext dataset, each containing 100 remote sensing images per category. Table 2 presents the classes in RSketch, Rsketch_Ext, Earth on Canvas, and UCMerge Landuse dataset. Figure 3 illustrates the sample data corresponding to each category in this dataset. GF-1 images tiles specifically depicted a section of the Yangtze River in Anhui Province, China. There were 4842 image tiles, and each image tile was formatted to a resolution of 256×256 pixels.

**Table 2.** Classes in RSketch, RSketch_Ext, Earth On Canvas and UCMerge Landuse dataset.

| RSketch and RSketch_Ext | Earth on Canvas | UCMerge Landuse |
|---|---|---|
| Airplane | Airplane | Airplane |
| Baseball Diamond | Baseball Diamond | Baseball Diamond |
| Golf Course | Golf Course | Golf Course |
| Intersection | Intersection | Intersection |
| Overpass | Overpass | Overpass |
| River | River | River |
| Runway | Runway | Runway |
| Storage Tanks | Storage Tanks | Storage Tanks |
| Tennis Court | Tennis Court | Tennis Court |
| Basketball Field | Buildings | Buildings |
| Beach | Freeway | Beach |
| Bridge | Harbor | Agricultural |
| Closed Road | Mobile Home Park | Chaparral |
| Crosswalk | Parking Lot | Dense Residential |
| Football Field | | Forest |

|  |  |
|---|---|
| Oil Gas Field | Freeway |
| Railway | Harbor |
| Runway Marking | Medium Residential |
| Swimming Pool | Mobile Home Park |
| Wastewater Treatment Plant (WWTP) | Parking Lot |
|  | Sparse Residential |

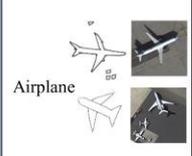

**Figure 3.** Samples in the RSketch_Ext dataset (presenting two sketches and two remote sensing images for each category, one sketch is from original RSketch and one sketch is from our extension).

In this study, we implemented a 4-fold cross-validation approach to evaluate our model. The 20 categories from both the RSketch and RSketch_Ext datasets were divided into four distinct sets of classes. These sets were designated as S1-S4. In each fold, 15 categories are assigned as seen classes and participate in training process, while the remaining 5 categories were reserved as unseen classes, used exclusively for testing. The four sets of unseen classes are detailed in Table 3. Within each seen class, we adopted a split in each set whereby 50% of the remote sensing images were utilized for training, and the remaining 50% were reserved for testing purposes. This partitioning provides a balanced approach for model training and evaluating, ensuring a comprehensive assessment across a diverse range of categories.

**Table 3.** Unseen classes in each fold.

| S1 | S2 | S3 | S4 |
|---|---|---|---|
| airplane | baseball diamond | basketball court | beach |
| bridge | closed road | crosswalk | football field |
| golf course | intersection | oil gas field | overpass |
| railway | river | runway | runway marking |
| storage tank | swimming pool | tennis court | WWTP |

*4.2. Implementation Details*

In this research, the proposed method was implemented using PyTorch framework on a single NVIDIA GeForce RTX 3080Ti GPU. Regarding the preprocessing of sketches and remote sensing images for model training, several steps of preprocessing were undertaken. All images were cropped according to the bounding box of strokes and then uniformly scaled to a resolution of 224×224 pixels, conforming to the requirements of the pre-trained model used in this study. This preprocessing step was crucial to eliminate redundant information and allowed the self-attention stage to focus more effectively on pertinent feature information.

The kernel size of convolutional layers used in the multi-level feature extraction stage was set at 7×7 for the initial convolutional layer, while for the subsequent three layers at 3×3. The stride parameter for all these convolutional layers was uniformly maintained at 2. This configuration resulted in the generation of 196 visual tokens, with each token represented as a 768-dimensional vector. In terms of architecture, the self-attention stage blocks were designed akin to those in the Vision Transformer (ViT), comprising 12 blocks. These blocks were pre-trained on the ImageNet-1K dataset [68]. The cross-modal attention stage had a single layer with 12 heads. During training process, the AdamW [69] optimizer was employed, with the learning rate set to $2×10^{-5}$ to balance efficient training convergence with the need for network stability.

*4.3. Evaluation criteria*

The evaluation criteria used in this study are:

(1) Mean Average Precision (mAP): This metric is a measure of retrieval accuracy, reflecting the mean area underneath the precision-recall curve in multiple queries.

(2) Top-K Accuracy：This metric measures the proportion of correctly retrieved images within the top K results returned by the model. In this research, we specifically evaluate the model's performance at three levels: the top 10, top 50, and top 100 retrieved images. These varying levels provide a comprehensive assessment of the model's retrieval accuracy and its ability to rank relevant images effectively.

Both mAP and Top-K Accuracy are critical in evaluating the model's performance in retrieving relevant images from the dataset, offering a multi-dimensional understanding of its effectiveness in various test scenarios.

*4.4. Experiments on RSketch dataset*

To verify the effectiveness of the proposed method in this study, we compared the proposed model with the following baseline deep learning methods for cross-modal retrieval: MR-SBIR [8], DAL [19], DSM [70], DOODLE [71], DSCMR [72], CMCL [73], and ACNet [18].

Among the benchmark methods considered, only the initial baselines are specifically tailored for Sketch-Based Remote Sensing Image Retrieval (SBRSIR). Due to the limited availability of methods dedicated to SBRSIR, we have included additional methodologies from the broader field of computer vision, which are typically applied to conventional images. Comparing our specialized method against these more generalized SBIR approaches may provide insights into the feasibility of adapting algorithms designed for ordinary images to the remote sensing context.

In this study, both the proposed method and its counterparts were initially trained and tested using the RSketch dataset as a benchmark. The outcomes for each network were obtained by averaging the results across the four folds, as detailed in Table 4.

**Table 4.** Test result (%) with RSketch dataset.

|  | Seen | | | | Unseen | | | |
|---|---|---|---|---|---|---|---|---|
|  | **mAP** | **Top10** | **Top50** | **Top100** | **mAP** | **Top10** | **Top50** | **Top100** |
| MR-SBIR | 83.75 | 88.77 | 85.59 | 77.20 | 47.86 | 57.70 | 49.96 | 42.44 |
| DAL | 92.56 | 94.97 | 93.81 | 89.19 | 50.67 | 61.90 | 54.62 | 46.02 |
| DSM | 56.80 | 71.60 | 64.06 | 55.00 | 19.29 | 21.90 | 21.06 | 19.71 |
| DOODLE | 49.11 | 56.67 | 48.33 | 23.67 | 33.24 | 35.00 | 33.00 | 31.50 |
| DSCMR | 96.12 | 96.37 | 96.82 | 94.80 | 46.60 | 27.00 | 49.26 | 41.71 |
| CMCL | 95.14 | 96.07 | 95.59 | 93.41 | 41.61 | 51.30 | 43.20 | 36.50 |
| ACNet | 38.11 | 44.20 | 36.54 | 32.44 | 25.14 | 28.10 | 23.73 | 21.32 |
| **Ours** | **98.17** | **98.50** | **98.31** | **95.93** | **76.62** | **85.00** | **79.96** | **70.01** |

The data in Table 4 reveals that our proposed method attained a mean Average Precision (mAP) of 98.17% and a Top-100 accuracy of 95.93% for seen classes. Notably, the performance gap between our model and the baseline methods was more pronounced in unseen classes, with the proposed method achieving a mAP of 76.62% and a Top-100 accuracy of 70.01%. These results suggest that the method presented in this study significantly surpassed other baseline methods in terms of remote sensing image retrieval accuracy. Also, the remote sensing specific methods performed better than DOODLE, DSCMR and ACNet, which is designed for ordinary images, highlighting the domain gap between the retrieval of ordinary images and remote sensing images. The proposed method also achieved equivalent or slightly better mAP compared with the latest CBIR algorithms [5]. We detected no result difference between TIFF and JPEG encoded sketches.

We conducted a systematic investigation to determine how various hyperparameters influence the performance of the model. One of the key parameters adjusted was the number of heads in the multi-head attention mechanism. We observed that increasing the head count from 8 to 12 led to a notable improvement of 1.8% in the model's mean Average Precision, and further increasing it from 12 to 16 led to a small additional improvement of 0.9% in mAP. Conversely, a reduction in the number of heads from 8 to 4 resulted in a mAP decrease of 2.2%. This finding suggests a direct correlation between the number of attention heads in the attention mechanism and the accuracy of retrieval. Increasing the number of attention heads appropriately can further improve the accuracy of retrieval. However, it should also be noted that too many attention heads will increase the complexity and computational cost of the model and may lead to overfitting of the model, especially the GPU memory consumption during training, and may lead to overfitting of the model. In our test, the training time prolonged when the number of heads increased. The training time for networks with 12 attention heads was about 2.8% longer than those with 4 attention heads, and 16 attention heads required an additional 0.9% longer time. Therefore, the appropriate number of attention heads should be determined by considering model performance, complexity, computational efficiency and overfitting. Moreover, modifications to the learning rate of the AdamW optimizer also demonstrated significant impacts on the model's retrieval accuracy. An increase in the learning rate from $1\times10^{-5}$ to $2\times10^{-5}$ yielded a 0.6% increase in the mAP value. In contrast, reducing the learning rate from $1\times10^{-5}$ to $5\times10^{-6}$ led to a substantial decline in mAP, a decrease of 5.9%. These insights highlight the sensitivity of the model to specific hyperparameter settings, underscoring the importance of careful tuning to optimize retrieval performance.

*4.5. Experiments on RSketch_Ext dataset*

Evaluating our approach using the RSketch dataset facilitates a direct benchmarking against existing methodologies. Nonetheless, the RSketch dataset presents considerable limitations in terms of category breadth, dataset size, and diversity. To more accurately assess the effectiveness and efficiency of our proposed method, we expanded our experimental framework to include the RSketch_Ext dataset, allowing for a more comprehensive evaluation. Compared to RSketch, this dataset maintains the same number of categories but with a substantially increased quantity of images and sketches in each. Notably, these additional images and sketches are sourced from a variety of different datasets and creators, providing a broader range of data inputs in both training and testing.

While keeping other settings unchanged, we trained the proposed model and three baseline models on the expanded RSketch_Ext dataset. Additionally, we also utilized the model trained on the RSketch dataset mentioned in this paper (marked Ours-RSketch). A total of five models were tested on the expanded RSketch_Ext dataset, and the results of the tests are shown in Table 5. This comparison aims to shed light on the relationship between data characteristics – particularly in terms of volume and source diversity – and the model's efficacy.

**Table 5.** Model test results (%) on RSketch_Ext dataset.

|  | Seen | | | | Unseen | | | |
|---|---|---|---|---|---|---|---|---|
|  | **mAP** | **Top10** | **Top50** | **Top100** | **mAP** | **Top10** | **Top50** | **Top100** |
| MR-SBIR | 75.45 | 87.60 | 85.15 | 82.86 | 35.23 | 53.60 | 46.70 | 42.13 |
| DOODLE | 48.43 | 54.39 | 46.85 | 25.47 | 30.16 | 32.20 | 26.57 | 22.50 |
| ACNet | 40.90 | 37.60 | 34.40 | 30.40 | 29.08 | 31.60 | 28.72 | 26.40 |
| Ours | 97.88 | 98.73 | 99.05 | 99.07 | 70.79 | 82.20 | 80.04 | 77.80 |
| Ours-RSketch | 98.17 | 98.50 | 98.31 | 95.93 | 76.62 | 85.00 | 79.96 | 70.01 |

From Table 5, it can be observed that the retrieval performance of the proposed model outperforms other baseline methods. Also, the model trained and tested using the extended dataset showed a little change in retrieval accuracy for seen classes compared to the model trained on the original RSketch dataset. This is mainly because the model trained on the original dataset already achieved high accuracy for seen classes, so there is not much room for improvement in retrieval accuracy. However, in the retrieval of unseen classes, the model trained on the expanded dataset exhibits a decrease of 6% in mAP value and a decrease of approximately 3% in Top-10 accuracy compared to the model trained on the original dataset. This result may be attributed to the expansion of the training datasets, with an increase in the diversity of remote sensing sources per category, the difficulty of the model to catch feature patterns also increases, leading to a decrease in mAP value for unseen classes. Meanwhile, the values for Top-50 showed minimal changes, while there was an increase of around 7% in Top-100 accuracy. The discrepancy between the performance change in mAP, Top10 and Top100, saying better mAP and Top10 comes with worse Top100, is possibly a common phenomenon in such a test, as the results of DOODLE and ACNet demonstrate a similar pattern. A possible explanation is that the more sensible model may miss some potential candidate within the borderline, resulting worse Top100. Such result may indicate an inevitable tradeoff between the model's capability for retrieving the most viable candidate in the first few attempts and the capability for catching more candidate in a larger candidate pool.

In addition to quantitative analysis, we conducted a manual examination of the retrieval results. Figures 4 and 5 display selected results of our model for both seen and unseen classes within the RSketch_Ext dataset, where the proposed method achieves high accuracy in retrieving seen classes. Notably, in certain unseen categories such as basketball courts and runway, the model also demonstrates high correct retrieval rates. Although there were two instances of incorrect retrievals in the crosswalk category, a closer inspection revealed that the incorrectly retrieved images did indeed feature crosswalk, albeit categorized under "intersection" in the test dataset. While the retrieval results for the oil and gas field category were not accurate, there was a noticeable similarity in texture and structure between the sketches and the retrieved remote sensing images.

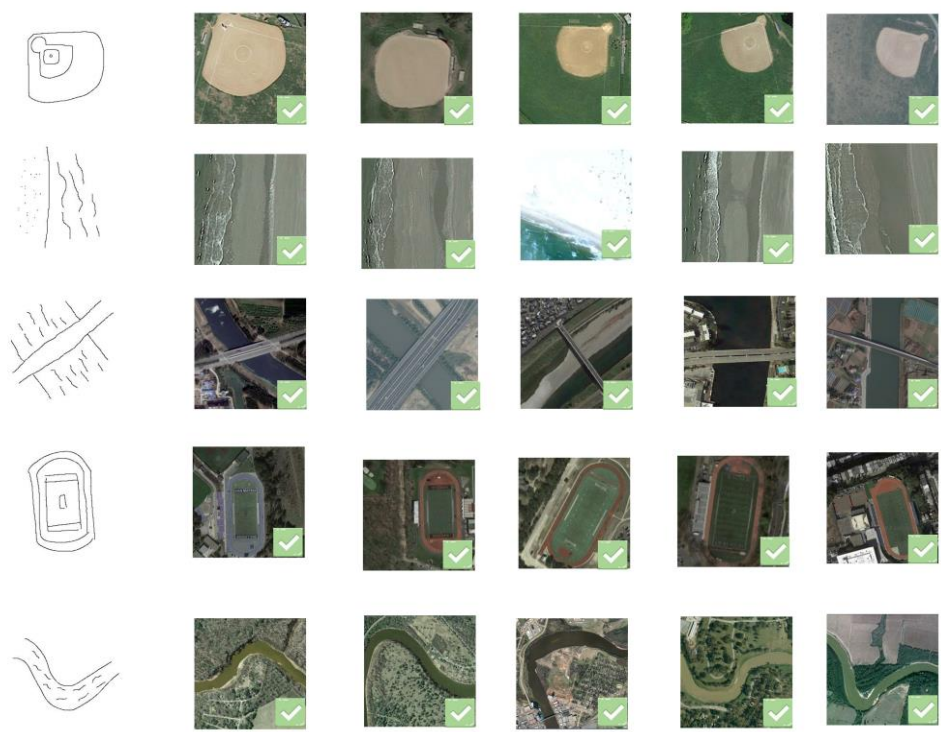

**Figure 4.** Retrieval results of 5 seen categories in a test on RSketch_Ext dataset with the top 5 retrieval results. The green checkmark in the lower right corner of the picture marks the correct retrieval, and the red X in the lower right corner marks the wrong retrieval. The five categories selected are baseball diamond, beach, bridge, football field, and river.

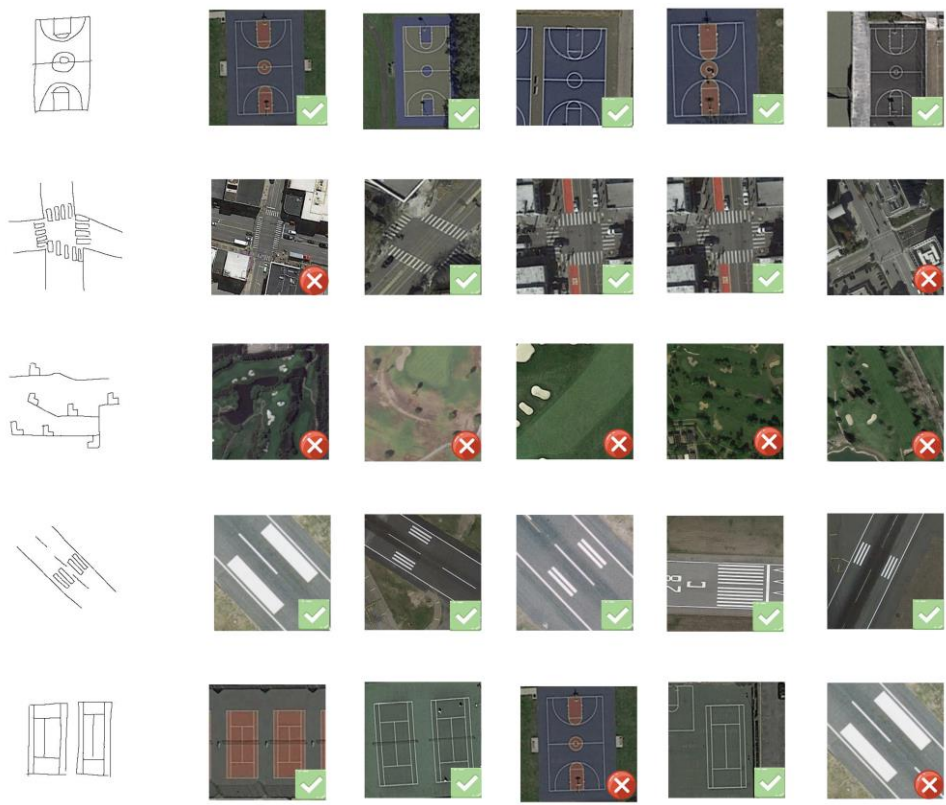

**Figure 5.** Retrieval results of 5 unseen categories in a test on RSketch_Ext dataset with the top 5 retrieval results. The green checkmark in the lower right corner of the picture marks the correct retrieval, and the red X in the lower right corner marks the wrong retrieval. The five categories selected are basketball court, crosswalk, oil gas field, runway, and tennis court.

*4.6. Experiments on Earth on Canvas and UCMerge Landuse datasets*

To assess the proposed model's zero-shot learning and domain adaptation capabilities, we designed an experiment in which our proposed model and three baselines, initially trained on the RSketch_Ext dataset, underwent testing on the distinct datasets of Earth on Canvas and UCMerge Landuse. For the UCMerge Landuse dataset, the experiment only incorporated data from the 10 categories that overlap with those in the RSketch_Ext dataset. The experiments were also categorized into seen and unseen classes, based on whether the classes being tested were included in the model's training dataset. The test results are detailed in Table 6 and Table 7. This approach of testing across differing datasets provides valuable insights into the model's ability to generalize beyond its training dataset and its zero-shot capability in such situation, offering a stringent test of its practical applicability in real-world scenarios.

**Table 6.** Model test results (%) on Earth On Canvas dataset.

|  | Seen | | | | Unseen | | | |
|---|---|---|---|---|---|---|---|---|
|  | mAP | Top10 | Top50 | Top100 | mAP | Top10 | Top50 | Top100 |
| MR-SBIR | 61.63 | 63.95 | 56.45 | 48.91 | 42.72 | 52.10 | 45.98 | 41.06 |
| DOODLE | 47.78 | 52.00 | 48.00 | 40.00 | 21.62 | 28.00 | 21.33 | 16.00 |
| ACNet | 34.87 | 30.80 | 31.68 | 29.32 | 30.96 | 26.40 | 23.84 | 24.28 |
| Ours | **83.31** | **85.61** | **82.56** | **76.83** | **63.32** | **71.10** | **64.16** | **56.98** |

**Table 7.** Model test results (%) on UCMerge Landuse dataset.

|  | Seen | | | | Unseen | | | |
|---|---|---|---|---|---|---|---|---|
|  | mAP | Top10 | Top50 | Top100 | mAP | Top10 | Top50 | Top100 |
| MR-SBIR | 81.01 | 88.37 | 82.47 | 71.93 | 85.54 | 85.17 | 79.47 | 71.73 |
| DOODLE | 57.03 | 73.33 | 54.67 | 30.67 | 37.23 | 40.00 | 28.00 | 24.00 |
| ACNet | 38.48 | 32.80 | 31.12 | 27.72 | 31.79 | 31.20 | 28.00 | 26.36 |
| Ours | **98.00** | **98.78** | **98.68** | **95.35** | **89.23** | **96.42** | **91.87** | **82.43** |

The results from Tables 6 and 7 indicate that our model, even when trained on the RSketch_Ext dataset, still demonstrates very good and superior performance compared to other baseline methods when tested on the Earth on Canvas dataset and the UCMerge Landuse dataset. These results demonstrated the robust domain adaptation ability of the proposed model, particularly in its capacity to effectively retrieve relevant data from a dataset comprising previously unseen data and classes. This underscored the model's adaptability and potential for practical application in diverse and novel retrieval scenarios.

*4.7. Experiments on GF-1 tiles*

In this part of the study, we further evaluated the practical application of the model trained on the RSketch_Ext dataset by conducting retrieval experiments using real-world remote sensing image tiles within the Habitat Yangtze project. As illustrated in Figure 6, the remote sensing image tiles utilized for these experiments were from the GF-1 satellite and there are a total of 4842 tiles, each with a size of 256×256 pixels. For the retrieval input, sketches from the RSketch_Ext dataset were employed. It is important to note that the GF-1 satellite imagery is not part of the training data. Consequently, this set of experiments is also a valuable test of the model's capabilities in handling completely unseen data. The use of real satellite imagery in these tests provided a stringent assessment of the model's retrieval accuracy and efficiency in realistic scenarios with remote sensing image tiles of a large research region, beyond the confines of the training dataset.

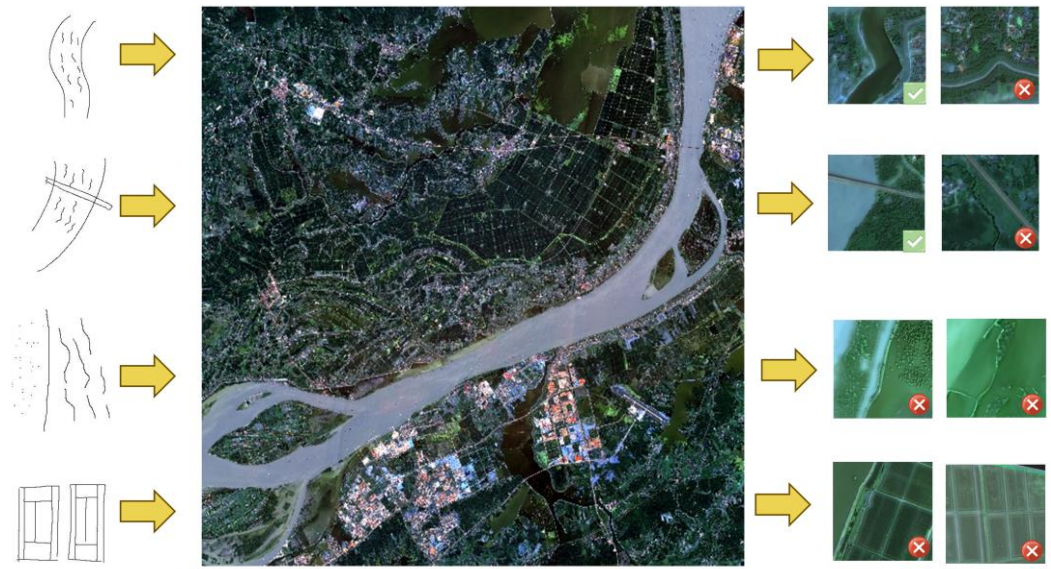

**Figure 6.** Retrieval results of GF-1 image tiles.

Figure 6 showcases the retrieval capabilities of the proposed model when applied to GF-1 remote sensing image tiles, particularly focusing on river and bridge classes. By manually examining the top 10 retrieved images for river and bridge categories, we observed an accuracy of 68% for the top 10 images in the river category and 70% for the top 10 images in the bridge category. This result demonstrated the effectiveness of our proposed model in a practical setting, the subjective feedback from the user is satisfied. Despite the model's proficiency in retrieving relevant images, it is noteworthy that some retrieved images, while not precisely matching the query's class, bear a resemblance in shape and texture to the input sketches. This is particularly evident in the retrieval attempts for the beach and tennis court classes, which are not presented in the selected research area. Although the images returned for these categories do not strictly belong to the specified classes, they share a similarity in shape with the sketches used for retrieval. We conducted retrieval tests on the same 100 query sketches respectively in the image tile database of GF-1 and the pre-calculated GF-1 retrieval tokens. With a same retrieval accuracy, the retrieval time in the GF-1 image tile database was 66.3 seconds, while the retrieval time in the pre-calculated retrieval tokens was only 11.7 seconds. From the comparison of the retrieval times of the two, it can be clearly seen that the retrieval time using pre-calculated retrieval tokens is much shorter than the time for retrieval using the original image tiles. This confirms that using pre-calculated retrieval tokens can effectively improve the efficiency of retrieval in practical applications of large-scale remote sensing image retrieval, and retrieval efficiency is one of the key factors in retrieving large-scale remote sensing image datasets.

This observation underlines a significant aspect of the proposed model's functionality: its ability to discern and match shapes between sketches and remote sensing images, exceeding mere categorical correspondence. This capability is particularly useful in instances where the category depicted in the queried sketch is absent from the dataset. In such cases, the model suggests alternative images that, while not categorically identical, are visually similar to the sketch in terms of shape. This outcome indicates the model's potential for broader applications, where shape recognition plays a crucial role in retrieval processes.

## 5. Discussion

The method introduced in this study, incorporating multi-level feature extraction and attention-guided tokenization, offers a novel deep learning approach for Sketch-Based Remote Sensing Image Retrieval. Our findings demonstrated the effectiveness of this method, which outperformed seven baseline methods, particularly under zero-shot conditions. Additionally, the method exhibited good domain adaptation capabilities. It could effectively handle remote sensing images from sources not included in the training dataset, thus underscoring its robustness and flexibility. A notable feature of the proposed method is the pre-computation of retrieval tokens for each remote sensing image in the database, enabling accelerated retrieval processes through vector search algorithms.

Despite ablation studies and the promising results from experiments across five datasets, further exploration of the network's full potential remains pertinent. For instance, the current approach utilizes multi-level tokenization for sketches and applies identical token filtering mechanisms for both sketches and remote sensing images, which could be optimized separately for each modality. Moreover, leveraging the latest pre-trained networks could enhance the self-attention model. Our current training process begins with a network pre-trained on the ImageNet-1K dataset. Exploring other cutting-edge pre-trained networks may yield additional improvements.

For experiment dataset, the study's expansion of the RSketch dataset to Rsketch_Ext focuses primarily on increasing the volume and diversity of sketches and remote sensing images per class, rather than expanding the range of scenario classes. Given that the current 20 classes are insufficient to encompass all potential scenarios and their semantic categories, a significant expansion of the training dataset's class diversity is necessary but poses challenges in terms of time and resources. An automated approach for extending the benchmark dataset would be invaluable for future research in this domain. We recognize the presence of several successful applications of Graph Convolutional Networks (GCNs) within the Sketch-Based Image Retrieval (SBIR) domain that leverage semantic information. Our proposed method has already demonstrated exceptional performance in terms of effectiveness and efficiency, even without the integration of GCNs. Being semantic knowledge independent is an advantage of our proposed method as it can greatly reduce the burden of training data construction. However, exploring and incorporating a GCN-based design presents a promising avenue for future research, particularly when extensive training datasets encompassing a broad spectrum of scene semantics become available.

The sketches used in this study were limited to uniform line strokes, yet sketches can vary widely in form, including variations in line width, texture depiction, and annotations. Investigating how to adapt our model to accommodate these diverse sketch forms presents an intriguing avenue for future research. At the same time, the quality or style of sketches can also affect the effectiveness of retrieval. Generally, when using sketches to retrieve remote sensing images, sketches that contain enough information and at the same time with simple outlines often achieve better result. Too many internal details of sketches may distract the model's attention and may impair the effectiveness of categorical remote sensing retrieval. Furthermore, the incorporation of sketch annotations could introduce a natural language modality, transitioning from current two-modality to a three-modality SBRSIR framework, the three-modal retrieval framework can be used to retrieve remote sensing images similar in shape to the sketch. Building upon this, more refined retrieval can be conducted using text or other semantic information, such as customized retrieval based on the color of the retrieved images or the spatial relationships of objects within the images. Our tests with GF-1 image tiles also suggest potential of our model for fine-grained retrieval, making further investigation and enhancement of the model for fine-grained remote sensing image retrieval plausible.

For the practical application of this method in real-world settings, it can be deployed by constructing a remote sensing image retrieval website with our trained model. Users can upload or draw sketches using a mouse on the website interface, which are then sent to the backend of the website. The deployed model is called in the backend for retrieval from the pre-calculated remote sensing image database, and the retrieval results are then returned to the user interface. Additionally, this retrieval method can be combined with traditional retrieval methods (such as by timestamp or image region) for joint retrieval.

**6. Conclusions**

In this paper, we introduce a novel categorical zero-shot sketch-based remote sensing image retrieval method, leveraging multi-level feature extraction, self-attention-guided tokenization and filtering, and cross-modality attention update. The efficacy of this method has been thoroughly evaluated through experiments conducted on five remote sensing datasets. The results clearly indicate that our method not only surpasses other baseline methods but also exhibits strong zero-shot learning and domain adaptation capabilities. Particularly noteworthy is the network's ability to retrieve remote sensing images from both unseen categories and unseen data sources. Also, our model is semantic knowledge independent, thus can greatly reduce the complexity of constructing training dataset compared with other method that leveraging semantic knowledge. Our model is especially relevant for large-scale remote sensing datasets, as it enables the pre-calculation of retrieval tokens for all images in a database, enhancing scalability. Another contribution of this research is the manual expansion of the RSketch dataset into the RSketch_Ext dataset. This expansion, which substantially increases both the volume and diversity of the dataset, provides valuable insights into the performance of SBRSIR algorithms. We have made both the code of our method and the RSketch_Ext dataset publicly available online. This initiative aims to enable and encourage ongoing research and innovation in the area of sketch-based remote sensing image retrieval.

**Author Contributions:** Conceptualization, C.W. and X.M.; methodology, B.Y. and C.W.; software, B.Y.; validation, B.Y. and B.S.; formal analysis, C.W.; resources, C.W.; data curation, B.S., Z.L. and F.S.; writing—original draft preparation, B.Y. and C.W.; writing—review and editing, B.Y., C.W. and F.S.; visualization, B.Y.; project administration, C.W. and X.M. All authors have read and agreed to the published version of the manuscript.

**Funding:** This work was supported in part by the National Natural Science Foundation of China under Grants 41901410. Natural Science Research Project of Anhui Educational Committee under Grants 2023AH050103.

**Data Availability Statement:** We present the codes openly available at https://github.com/Snowstormfly/Cross-modal-retrieval-MLAGT to encourage more extensive research and applications in the field of sketch-based remote sensing image retrieval.

**Acknowledgments:** We would like to thank Anhui Province Key Laboratory of Wetland Ecosystem Protection and Restoration, Anhui University to provide us the hardware necessary for this project. We also thank the Space Climate Observatory Habitat Yangtze project, as this model is one of the project's outcomes.

**Conflicts of Interest:** No potential conflict of interest was reported by the author(s).